\documentclass[10pt,twocolumn,letterpaper]{article}

\usepackage{iccv}
\usepackage{times}
\usepackage{epsfig}
\usepackage{graphicx}
\usepackage{amsmath}
\usepackage{amssymb}
\usepackage{bm}
\usepackage{breqn}
\usepackage{multirow}
\usepackage{cite}

\usepackage[breaklinks=true,bookmarks=false]{hyperref}

\iccvfinalcopy 

\def\iccvPaperID{} 

\ificcvfinal\fi

\begin{document}

\title{SFHarmony: Source Free Domain Adaptation for Distributed Neuroimaging Analysis}

\author{Nicola K Dinsdale$^1$\\
\and
Mark Jenkinson$^{2,3,4}$\\
\and 
Ana IL Namburete$^{1,2}$\\
\and
\small
1. Oxford Machine Learning in NeuroImaging (OMNI) Lab, Department of Computer Science, University of Oxford, UK \\
\small
2. Wellcome Centre for Integrative Neuroimaging, FMRIB, University of Oxford, Oxford, UK \\
\small
3. Australian Institute for Machine Learning (AIML), Department of Computer Science, University of Adelaide, Adelaide, Australia \\
\small
4. South Australian Health and Medical Research Institute (SAHMRI), North
Terrace, Adelaide, Australia \\
\small
\url{nicola.dinsdale@cs.ox.ac.uk}
}

\maketitle
\ificcvfinal\thispagestyle{empty}\fi

\begin{abstract}
To represent the biological variability of clinical neuroimaging populations, it is vital to be able to combine data across scanners and studies. However, different MRI scanners produce images with different characteristics, resulting in a domain shift known as the `harmonisation problem'. Additionally, neuroimaging data is inherently personal in nature, leading to data privacy concerns when sharing the data. To overcome these barriers, we propose an Unsupervised Source-Free Domain Adaptation (SFDA) method, \texttt{SFHarmony}. Through modelling the imaging features as a Gaussian Mixture Model and minimising an adapted Bhattacharyya distance between the source and target features, we can create a model that performs well for the target data whilst having a shared feature representation across the data domains, without needing access to the source data for adaptation or target labels. We demonstrate the performance of our method on simulated and real domain shifts, showing that the approach is applicable to classification, segmentation and regression tasks, requiring no changes to the algorithm. Our method outperforms existing SFDA approaches across a range of realistic data scenarios, demonstrating the potential utility of our approach for MRI harmonisation and general SFDA problems.  Our code is available at \url{https://github.com/nkdinsdale/SFHarmony}.

\end{abstract}


\section{Introduction}
\begin{figure*}[h]
    \centering
    \includegraphics[width=0.95\textwidth]{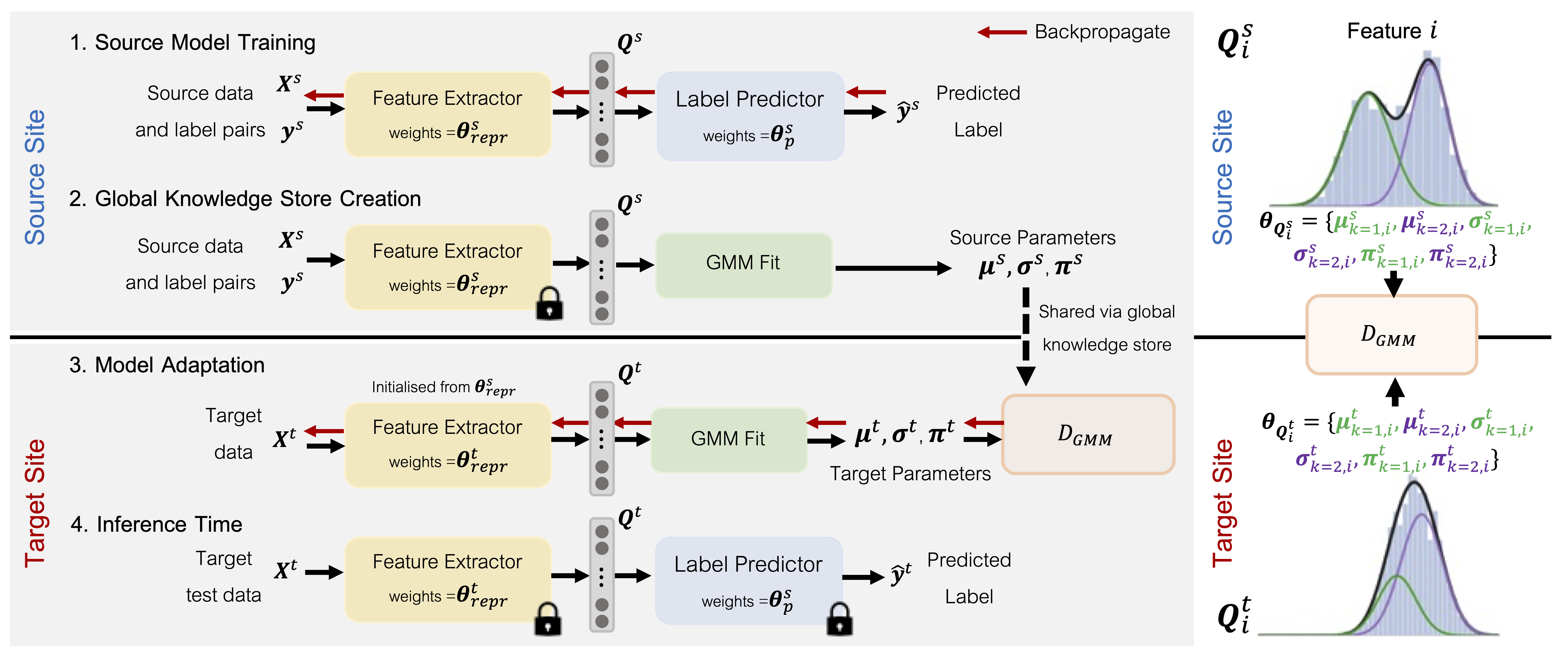}
    \caption{Schematic of the proposed \texttt{SFHarmony} method. The method fits a GMM to the source features, shares these via a global model store, and then completes SFDA by aligning the source and target feature distributions utilising a modified Bhattacharyya distance. $\bm{Q}^s$ is the source feature representation and $\bm{Q}^t$ is the target feature representation, and the figure shows the GMM setup, for a single feature $i$, when we are working with $K$, the number of components, being 2. This trivially generalises to more or less components.}
    \label{fig:method}
\end{figure*}

Deep learning (DL) models have proved to be powerful tools for neuroimage analysis. However, the majority of neuroimaging datasets remain small, posing a challenge for the training of sophisticated architectures with many  parameters. Thus, it is common practice to combine data from multiple sites and MRI scanners, both to increase the amount of data available for training, and to represent the breadth of biological variability that can be expected in diverse populations. However, the combination of data across MRI scanners with different acquisition protocols and hardware leads to an increase in non-biological variance \cite{Han2006,Jovicich2006,Takao2013}, which can be large enough to mask the biological signals of interest \cite{Takao2011}, even after careful pre-processing with state-of-the-art neuroimaging pipelines \cite{Glocker2019MachineLW}. The development of \textit{harmonisation} methods is therefore vital to enable the joint unbiased analysis of neuroimaging data from different scanners and studies.

The key goal for harmonisation methods is to be discriminative for the main task of interest whilst creating shared feature representations of the data across acquisition scanners, clearly mirroring the goal of domain adaptation (DA) \cite{Dinsdale2021}. The majority of deep learning based harmonisation methods are based on DA methods, either using adversarial approaches to create shared feature embeddings \cite{Dinsdale2021,Guan2021}, or using generative approaches to create harmonised images \cite{Dewey2019,Zuo2021}. 

However, the vast majority of existing methods fail to be applicable in many realistic data scenarios. For example,  MR images are inherently personal information so their sharing is protected by legislation, such as GDPR \cite{GDPR} and HIPAA \cite{HIPPA}. Thus, the assumption of centralised data stores for model training is infeasible, particularly when working with clinical imaging data, which will be essential in order to produce representative models \cite{Dinsdale2022,Varoquax2022}. Distributed learning offers a promising solution, but the few proposed distributed harmonisation methods \cite{Chen2022,Dinsdale2022b} assume the simultaneous presence of the source and target data. The source data may not be available for the adaptation phase, for instance, due to confidentiality agreements, loss of the source data, or computational constraints \cite{Bateson2022}. Further, federated DA methods such as \cite{Dinsdale2022b} would require retraining of the model to incorporate any new sites, which is infeasible and computationally expensive. 

Therefore, we explore an unsupervised DA setting where only the source model, instead of the source data, is provided to the unlabelled target domain for harmonisation, known as Source Free Domain Adaptation (SFDA). This setting inherently protects individual privacy, whilst allowing the efficient incorporation of new sites without requiring target labels. We propose a simple yet effective solution, termed \texttt{SFHarmony}, which aims to match feature embeddings from the source and target, through characterising the embeddings as a Gaussian Mixture model (GMM) and the use of a modified Bhattacharyya distance \cite{Bhattacharyya1946}. This requires no modifications to the training of the source model, and the only additional communication is of summary statistics of the source feature embedding, allowing it to be simply applied to existing architectures. The summary statistics contain no information about individuals. 

Our contributions are as follows: 1) We propose a new method for SFDA, \texttt{SFHarmony}, based on aligning feature embeddings, utilising a modified Bhattacharyya distance, requiring no changes to source training;  2) We demonstrate the method's applicability to classification, segmentation and regression tasks, and show that the approach outperforms existing SFDA methods for domain shifts experienced when working with neuroimaging data; 3) We demonstrate the robustness of the method to additional challenges likely to be faced when working with real world imaging data: differential privacy and label imbalance.

\section{Related Work}
\textbf{Unsupervised Domain Adaptation (UDA):} UDA aims to exploit the knowledge learned from a source dataset to help to create a discriminative model for a related but unlabelled target dataset \cite{Liang2020}. DL-based UDA approaches can broadly be split into three categories \cite{Liang2020}: discrepancy-based, reconstruction-based, and adversarial. Discrepancy based approaches aim to minimise a divergence criterion, which measures the distance between the source and target data distributions encoded in a learned feature space \cite{Courty2017,Kang2019,Long2015,Sun2016}. Reconstruction-based approaches, instead, use reconstruction as a proxy task to enable the learning of a shared representation for both image domains \cite{Bousmalis2016,Ghifary2016,Murez_2018_CVPR}. Finally, adversarial approaches deploy a discriminator  that aims to identify the source of the data; the model is trained both to do the task and to trick the discriminator, creating domain invariant features \cite{Ganin2015,Tzeng2015}. These methods all assume simultaneous access to the source and target data, which poses data privacy challenges. 

\textbf{Federated Learning and Domain Adaptation:} Federated learning (FL) has been proposed as a method to train models on distributed data \cite{McMahan2016}. The data are kept on their local servers, and users train local models with private data and communicate the weights or gradients between sites for aggregation. Many FL approaches focus on minimising the impact of distribution shifts between clients \cite{Karimireddy2020,Xie2021,FedProx}; however, most assume that the data at all sites are fully labelled. However,
federated DA enables the incorporation of an unlabelled site into the federation without sharing data. FADA \cite{Peng2020} is a federated DA method, where features are shared between sites in a global knowledge store. The sharing of features, however, still poses privacy concerns as images may be recoverable from the features \cite{Feng2021}. Thus, FedHarmony \cite{Dinsdale2022b} instead encodes the features as Gaussian distributions and thus only the mean and standard deviations of the features need to be shared. Both of these methods still assume access to the source data during training and rely on adversarial approaches that are often unstable and hard to train. Other federated DA methods produce domain-specific models or ensembles \cite{Feng2021,Peterson2019PrivateFL,Yang2020,Yao2021}, meaning that the final predictions depend on the domain of the data. 

\textbf{Source Free Domain Adaptation:} SFDA takes the federated approach a step further and assumes that there is no access to the source data available at all: only the source model is available for model adaptation. The majority of SFDA methods have been developed for classification \cite{Ding2022,Dong2021,Liang2020, Yang2021, Roy2022,Ahmed2021,Li2020,Xia2021}, with a few being proposed for segmentation \cite{Kundu2021,Kundu2022,Liu2021,Prabhu2021,Yang2022}. There are two main approaches taken for SFDA. The first set of approaches are generative, aiming to create source samples using the source model weights \cite{Liu2021,Li2020}. These approaches, however, pose concerns about individual privacy, especially when working with medical images and low numbers of samples \cite{Feng2021} and cannot be simply applied to complex target tasks, limiting their utility when working with MRI data \cite{Wang2022}. The second set aim to minimise model entropy to improve predictions, guided by various pseudo labelling or uncertainty techniques to prevent mode collapse \cite{Ding2022,Dong2021,Liang2020, Yang2021, Roy2022,Ahmed2021,Kim2021}. These methods are often effective, but largely limited to classification tasks, and may require changes to the source model training to be effective \cite{Liang2020, Yang2021}. AdaMI \cite{Bateson2022} was proposed directly for medical image segmentation, but requires an estimate of the proportion of each label to prevent mode collapse. This ratio is hard to estimate for labels with high variability across populations, such as tumours or lesions. We could not identify any methods proposed for regression, where the lack of softmax outputs limit the direct application of methods based on entropy minimisation. 

\textbf{Harmonisation:} Many existing harmonisation approaches are based on COMBAT \cite{Fortin2017,Pomponio2020,Chen2022b}, which uses a linear model to represent the scanner effects on image-derived features. DL-based approaches for harmonisation generally utilise a DA approach, with many being generative, aiming to produce `harmonised' images \cite{Cackowski2021,Dewey2019,Moyer2020,Zhao2019,Zuo2021}, while the other branch uses adversarial approaches to harmonise the learned model features for a given task \cite{Dinsdale2021,Guan2021}. All of these methods assume simultaneous access to the source and target data, with some even requiring paired data \cite{Dewey2019}. The only existing methods for harmonisation which consider data privacy are Distributed COMBAT \cite{Chen2022} and FedHarmony \cite{Dinsdale2022b}; however, both assume constant communication with the source site. 

\section{Method}
The aim of this work is to create a SFDA method applicable to neuroimaging tasks, and to demonstrate its suitability for MRI harmonisation. Thus, the goal is to create a model where two images with the same label would share a feature embedding, regardless of the acquisition scanner -- the domain of the data. We thus follow the framework of \cite{Tzeng2015} and consider the network to be formed of a feature extractor, with parameters $\bm{\Theta}_{repr}$, and a label predictor, $\bm{\Theta}_p$. This network architecture is the same across source and target sites. The general schematic for training is shown in Fig. \ref{fig:method}.

\subsection{Creation of the Source Model}
The first stage is the training of the source model. This assumes the availability of a labelled training dataset $D^s = \{\textbf{X}^s, \textbf{y}^s\}$, where the image and label pairs depend on the task of interest. Unlike some existing methods \cite{Liang2020,Yang2021}, our proposed approach requires no changes to the training of the source model or to the architecture. The model can thus be flexibly trained following the standard training procedure for the source data, with the goal being to create a well-trained source model. In our experiments, we consider the simplest source training, minimising a loss function ($L_{task}$) dependent on the task of interest with full supervision:
\begin{equation}
    L(\textbf{X}^s, \textbf{y}^s; \bm{\Theta}_{repr}^s, \bm{\Theta}_{p}^s) =  \frac{1}{N_s} \sum^{N_s}_i L_{task}(\textbf{X}_{i}^s, \textbf{y}_{i}^s)
\end{equation}
where $N_s$ is the total amount of labelled source data.

\subsection{Global Information Store}
For successful SFDA, we need to align the learned feature embedding, $\textbf{Q}^s = f(\textbf{X}^s, \bm{\Theta}_{repr}^s)$ for the source and target data. To achieve this without requiring the source data, we propose to follow the precedent of existing privacy-preserving medical imaging approaches \cite{Chen2022, Dinsdale2022b, Dong2021} and, thus, create a global knowledge store to share summary statistics of the features. In \cite{Dinsdale2022b}, it is proposed that the features can be encoded as Gaussian distributions, and thus the statistics to be shared would be a mean and standard deviation per feature. We hypothesise that for many tasks, especially classification tasks with discrete categories, simple Gaussian distributions are unlikely to sufficiently characterise $\textbf{Q}^s$. We thus propose to describe the features using a Gaussian mixture model (GMM), with each feature being encoded as an independent 1D GMM, such that, for feature $i \in {N_{Q^s}}$, where ${N_{Q^s}}$ is the number of features in ${\textbf{Q}^s}$:
\begin{equation}
    \textbf{Q}_{i}^s \sim \sum^K_{k=1} \bm{\pi}^s_{k,i} \mathcal{N}(\bm{X}^s; \bm{\mu}^s_{k,i}, \bm{\sigma}^{s^2}_{k,i})
\end{equation}
where $K$ is the number of components in the GMM, $\bm{\mu}^s_{k,i}$ and  $\bm{\sigma}^{s^2}_{k,i}$ are the mean and variance defining the $k^{th}$ Gaussian component of the $i^{th}$ feature for the source site, and $\bm{\pi}^s_{k,i}$ is the weighting factor for this $k^{th}$ Gaussian (which sum to one across components). Note that the features are considered before the activation function. The same number of components, $K$, are fit for all features. 

Thus, the GMM for feature $i$ is defined by the parameters:

\begin{equation}
    \bm{\Theta}^s_i = \{\bm{\pi}^s_{k,i}, \bm{\mu}^s_{k,i}, \bm{\sigma}^{s^2}_{k,i}\}, k=1..K
\end{equation}
and these parameters can be determined using Expectation Maximisation (EM), by finding the maximum likelihood estimate (MLE) of the unknown parameters:

\begin{equation}
    \mathcal{L}(\Theta_i) = \sum_{n=1}^{N_{s,i}} \log(\sum^K_{k=1} \bm{\pi}^s_{k,i} \mathcal{N}(\bm{X}^s_n; \bm{\mu}^s_{k,i}, \bm{\sigma}^{s^2}_{k,i}))
\end{equation}

\noindent for each feature $i$ in $\bm{Q^s}$. This, therefore, produces three parameter arrays that fully define the GMMs of the source features, which are communicated alongside the source weights to target sites:
\begin{equation}
    {\bm{\Theta}_{Q^s}} = \{\bm{\mu}^
s\in\mathbb{R}^{K \times {N_{Q^s}}}; \bm{\sigma}^{s^2}\in\mathbb{R}^{K \times {N_{Q^s}}}; \bm{\pi}^s\in\mathbb{R}^{K \times {N_{Q^s}}}\}.
\end{equation}
These parameters contain no individually identifying information, as they represent aggregate statistics across the whole population.
\subsection{Target Model Adaptation}
Given that we now have a well trained source model, with parameters $\bm{\Theta}_{repr}^s$ and $\bm{\Theta}_{p}^s$, and the source GMM parameters, $\bm{\Theta}_{Q^s}$, we can now adapt the model at any target site. We assume access to an unsupervised target, with only data samples $\bm{X}^t$ and no labels available. 

We initialise the target model using the source trained weights. Model adaptation only involves finetuning the feature extractor to match the learned feature distribution across the two sites. In adversarial approaches, a discriminator is added to the overall architecture that aims to distinguish between source and target samples. We could utilise this approach, following \cite{Dinsdale2022b}, by drawing feature samples, using the source GMM parameters  $\bm{\Theta}_{Q^s}$, but adversarial approaches are notoriously unstable and difficult to train. We therefore, instead, propose to minimise the difference between the source feature distribution and target feature distribution using the GMM parameters directly. 

Therefore, the first step of model adaptation is to calculate the current target features, $\textbf{Q}^t = f(\textbf{X}^t, \bm{\Theta}_{repr}^t)$, and then, using the same EM approach as above, we can create the parameters of the target GMM fit:
\begin{equation}
    {\bm{\Theta}_{Q^t}} = \{\bm{\mu}^t\in\mathbb{R}^{K \times {N_{Q^t}}}; \bm{\sigma}^{t^2}\in\mathbb{R}^{K \times {N_{Q^t}}}; \bm{\pi}^t\in\mathbb{R}^{K \times {N_{Q^t}}}\}.
\end{equation}

We propose to use a modified Bhattacharyya distance \cite{Bhattacharyya1946} as the loss function. The Bhattacharyya distance measures the similarity of two probability distributions, which for continuous probability distributions is defined as:
\begin{equation}
    D_B(p, q)  = - \ln (BC(p,q))
\end{equation}
where
\begin{equation}
    BC(p, q)  = \int_x \sqrt{p(x)q(x)} dx.
\end{equation}
The Bhattacharyya distance has a simple closed form solution when the two probability distributions are both Gaussian. If $p \sim \mathcal{N}({\mu_p, \sigma_p^2})$ and $q \sim \mathcal{N}({\mu_q, \sigma_q^2})$ then: 
\begin{equation}
    D_B(p, q) = \frac{1}{4} \frac{(\mu_p-\mu_q)^2}{\sigma_p^2+\sigma_q^2} + \frac{1}{2} \ln\Bigl(\frac{\sigma_p^2+\sigma_q^2}{2\sigma_p\sigma_q}\Bigl).
\end{equation}
There is, however, no equivalent closed form solution for a GMM. In \cite{Sfikas2005} they propose an approximation for the GMM as a sum of the Bhattacharyya distances for each pair of Gaussians in the mixture model, weighted by the associated $\bm{\pi}$ values. We suggest that this is not the most appropriate reformulation: we are more interested in the corresponding pairs of Gaussians than in the cross-relationships, as we do not wish to minimise the difference between cross pairs. Rather, we wish specifically to make the target distribution match the source. Thus, if we consider our target and source GMM distributions, parameterised by ${\bm{\Theta}_{Q^s}}$ and ${\bm{\Theta}_{Q^t}}$, we propose to use the following approximation:
\begin{multline}
    D_{GMM}({\bm{\Theta}_{Q^s}},{\bm{\Theta}_{Q^t}})  = \\
    \sum^M_{k=1} \bm{\pi}_{k}^s\bm{\pi}_{k}^t \Bigl(\frac{1}{4} \frac{(\bm{\mu}_{k}^s - \bm{\mu}_{k}^t)^2}{\bm{\sigma}_{k}^{s^2} + \bm{\sigma}_{k}^{t^2}} + \frac{1}{2} \ln \Bigl( \frac{\bm{\sigma}_{k}^{s^2} + \bm{\sigma}_{k}^{t^2}}{2 \bm{\sigma}_{k}^s\bm{\sigma}_{k}^t}  \Bigl) \Bigl)
\end{multline}
such that we find the weighted sum of the Bhattacharyya distances between each corresponding pair of Gaussians, where $k$ is the component in the GMM. It can further be seen that this approximation retains the desirable property that when ${\bm{\Theta}_{Q^s}}={\bm{\Theta}_{Q^t}}$, then $D_{GMM}({\bm{\Theta}_{Q^s}}, {\bm{\Theta}_{Q^t}}) = 0$. The correspondence between Gaussians can be ensured by simply ordering the parameters by the mean estimates.

\begin{table*}[h]
\begin{center}
\resizebox{0.95\textwidth}{!}{%
\begin{tabular}{|l|ccc|c|ccc|}
\hline
\multicolumn{1}{|c|}{\multirow{2}{*}{Method}} & \multicolumn{1}{c|}{\multirow{2}{*}{S}} & \multicolumn{1}{c|}{\multirow{2}{*}{T}} & \multirow{2}{*}{C}        & \multirow{2}{*}{\begin{tabular}[c]{@{}c@{}}Information \\ Communicated\end{tabular}} & \multicolumn{3}{c|}{Average Accuracy}                                                \\ \cline{6-8} 
\multicolumn{1}{|c|}{}                        & \multicolumn{1}{c|}{}                   & \multicolumn{1}{c|}{}                   &                           &                                                                                      & \multicolumn{1}{l|}{Batchsize 5} & \multicolumn{1}{l|}{Batchsize 50} & Batchsize 500 \\ \hline
Source Model & \checkmark & x  & x & -  & & 80.71 &
\\
\hline
Centralised Data & \checkmark  & \checkmark  & \checkmark & All Data & 88.34  & 91.65  & 91.27  \\
Target Finetune & x  & \checkmark  & \checkmark & Model Weights  & 88.15  & 88.96  & 83.26 \\ \hline
DeepCORAL \cite{Sun2016} & \checkmark & x & \checkmark & All Data & 82.43 & 83.85 & 83.65     \\
FADA \cite{Peng2020} & \checkmark & x & x & Model Weights + Features & 81.69 & 76.77 & 76.53\\
FedHarmony \cite{Dinsdale2022b} & \checkmark & x & x & Model Weights + Statistics & 81.48 & 76.12 & 76.20 \\
\hline
Minimise Entropy & x & x & x & Model Weights & 42.59 & 83.54 & 83.96 \\
SHOT \cite{Liang2020} (no smoothing) & x & x & x & Model Weights & 66.86 & 83.60 & 85.40 \\
SHOT \cite{Liang2020} (Source batchsize 5) & x & x & x & Model Weights & 72.06 & 74.44 & 74.61 \\
SHOT \cite{Liang2020} (Source batchsize 500) & x & x & x & Model Weights & 83.10 & 84.68 & 85.27\\
gSFDA \cite{Yang2021}  & x & x & x & Model Weights & 60.57 & 85.87 & 84.67 \\
USFAN \cite{Roy2022}  & x & x & x & Model Weights & 26.94 & 79.83 & 83.79 \\
\hline
SFHarmony 1 GMM Component  & x & x & x & Model Weights + Statistics & 85.47 & 85.71 & 86.16 \\
\hspace{0.5cm} w/o EM (Direct Fit) & x & x & x & Model Weights + Statistics & 77.26 & 76.99 & 86.03 \\
\hspace{0.5cm} w/o Batch Memory & x & x & x & Model Weights + Statistics & 80.25 & 82.13 & 84.60 \\
SFHarmony 2 GMM Components  & x & x & x & Model Weights + Statistics & \textbf{86.22} & \textbf{86.25} & \textbf{86.21} \\
SFHarmony 3 GMM Components  & x & x & x & Model Weights + Statistics &  86.21 & 85.70 & 85.96 \\
\hline

\end{tabular}%
}
\caption{Results on the OrganAMNIST classification task. S = Source data required, T = Target labels required, C = Centralised data. The average accuracy is across all 5 sites, weighted equally, and is reported for training batchsizes of 5, 50 and 500. Best SFDA method for each batchsize is in bold, other methods are included for reference. The w/o (without) components form an ablation study.}
\label{tab: mednist results}
\end{center}
\end{table*}

Thus, the feature extractor is finetuned for the target site by minimising $D_{GMM}$ averaged across all of the features in $\bm{Q}^s$. However, for each training iteration, only a fixed size batch is available to estimate the parameters, and for neuroimaging applications the maximum batchsize achievable is often small due to the relatively large image size \cite{Dinsdale2022}, which affects the estimate of the GMM parameters. As EM is sensitive to initialisation, to mitigate the small batch effect, we initialise the EM algorithm only once per training epoch, using the previous batch estimate as the initialisation for the next, providing memory between batches. The EM algorithm is reinitialised for validation (needed to calculate validation loss), preventing data leakage.

\subsection{Inference Time}
Finally, inference for the test data simply involves combining the finetuned feature encoder, $\bm{\Theta}_{repr}^t$, and the frozen source label predictor, $\bm{\Theta}_{p}^s$, such that $\hat{\bm{y}}^t = f(\bm{X}^t, \bm{\Theta}_{repr}^t, \bm{\Theta}_{p}^s)$. This therefore ensures that, given data from the source or target domain with the same feature embedding, the same label prediction is achieved across sites.

\section{Experimental Results}
To validate the effectiveness of our SFDA framework, we conduct a range of experiments with both simulated data with known domain shifts and real multisite MRI datasets, and we demonstrate the applicability of the method to classification, segmentation and regression tasks. 

\subsection{Datasets:}
 Further details for each dataset and model architectures are available in the Supplementary Materials.
 
 \textbf{OrganAMNIST} \cite{Xu2019} (Classification): curated as part of MedMNIST \cite{Yang2023}, we use OrganAMNIST as a test dataset. All images were pre-processed to 28 × 28 (2D) with the corresponding classification labels for 11 classes. We created simulated known domain shifts, to enable exploration of the method, with the strength of each shift designed to be such that a degradation in performance was seen across the sites.  The dataset was split into 5 sets, each with 5000 samples for training and 2000 for testing and the following domain shifts applied: 1) no shift (source site), 2) decreased intensity range, 3) increased intensity range, 4) Gaussian blurring, 5) salt and pepper noise, to model shifts likely across imaging sites. The backbone architecture took the form of a small VGG-like classifier, with categorical crossentropy as the task loss. Code to reproduce the data is provided. 

\textbf{CC359} \cite{Souza2018} (Segmentation): The dataset consists of brain images of healthy adults (29-80 years) acquired on MRI scanners from three vendors: Siemens, Philips and GE, at both 1.5 and 3T, with approximately 60 subjects per vendor and magnetic field strength. A 2D UNet was trained on slices from each site, then the performance when applied to the remaining sites was compared. As a result, the Phillips 1.5T was chosen as the source site as it had the largest performance drop. No additional preprocessing was applied to the images  apart from image resizing so that each subject volume was $128\times240\times160$. The data were split at the subject level per site, such that 40 subjects were available for training and 20 for testing. The segmentation task was skull stripping, using masks from the original study, and Dice loss was used as the task loss function. Further details and example segmentation masks can be found in the Supplementary Material.

\textbf{ABIDE} \cite{ABIDE} (Segmentation and Regression): Four sites (Trinity, NYU, UCLA, Yale) were used, so as to span age distributions and subject numbers. The data were split into training/test sets as 80\%/20\%, yielding a maximum of 127 subjects
for training (NYU) and a minimum of 35 (Trinity). NYU was the largest site, spanning the age distribution of all of the other sites, and so was chosen as the source site. For segmentation, we considered tissue segmentation (grey matter (GM), white matter (WM), CSF), using labels automatically generated using FSL ANAT. We used a 2D UNet trained on slices with Dice as the main task loss function. Dice score was averaged across the three tissues. For age prediction, a separate network was trained, following the setup and architecture in  \cite{Dinsdale2022b}, with MSE as the main task loss. Further details and example labels can be found in the Supplementary Material.

\textbf{Implementation Details:} All comparison methods used the same task-specific backbone architecture as the proposed method. Features were extracted in the second-to-last layer, before the activation function. Model architectures were chosen to give good source performance while allowing the use of large batchsizes, but most standard architectures could be used. Training was completed on an A10 GPU, using PyTorch 1.12.0. All models were trained with five-fold cross validation and results are presented on the hold out test set. A learning rate of $1\times10^{-6}$ was used for all datasets for adaptation with an AdamW optimiser.
\begin{figure}
\centering
\includegraphics[width=0.38\textwidth]{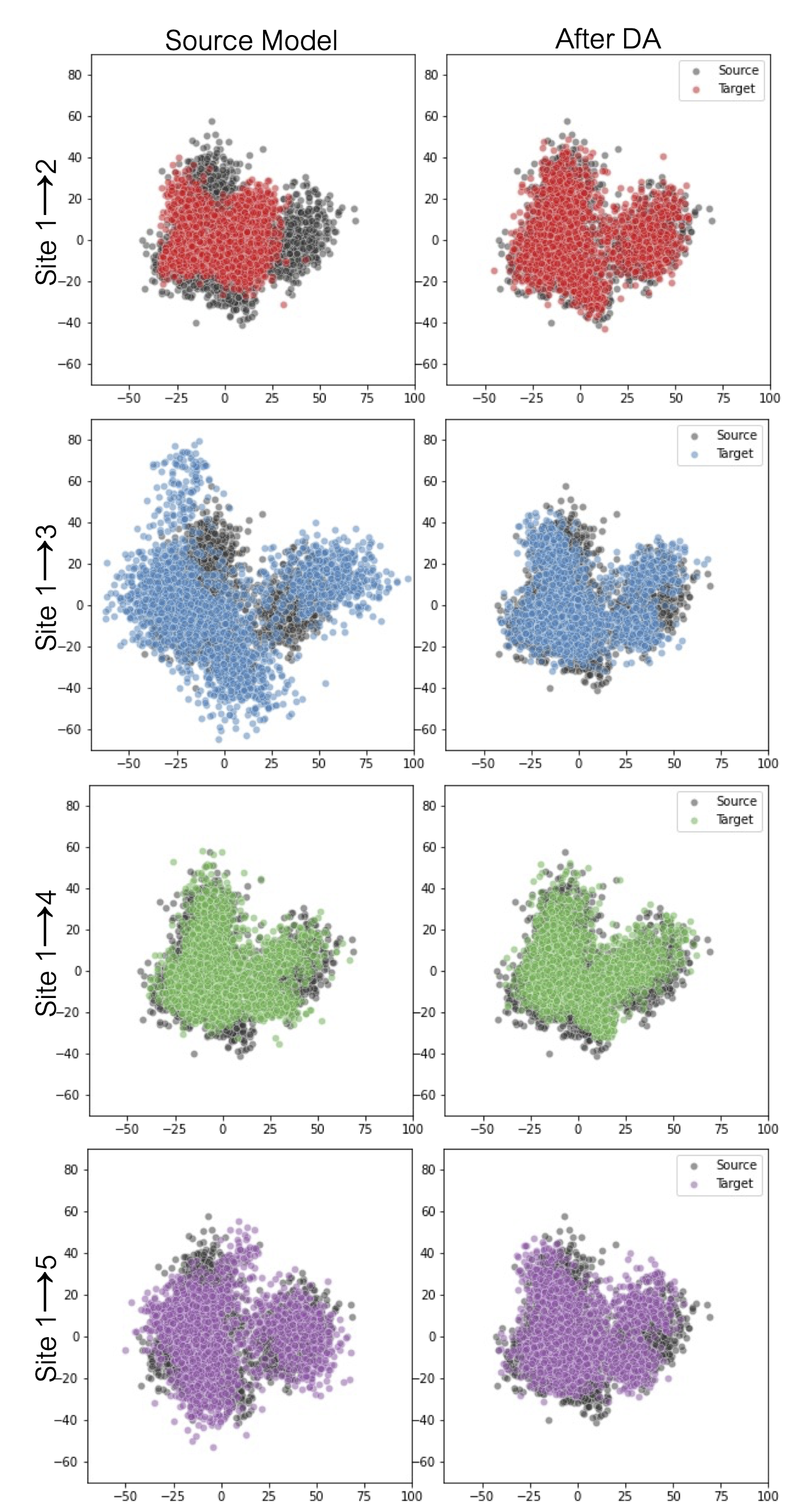}
\caption{PCA of $\bm{Q}_s$ and $\bm{Q}_t$ for each target site, before and after domain adaptation for the OrganAMNIST data, with simulated domain shifts. Black dots are the source features which are fixed and the colour represents the features for the relevant site. (Best viewed in colour.)}
\label{fig:features}
\end{figure}

\subsection{Classification: OrganAMNIST}
\textbf{Baselines:}
For the classification task, we first compare our approach to supervised oracles: source model only, centralised data, and target finetuning with frozen label predictor. We then compare to DeepCORAL \cite{Sun2016}, and two federated DA approaches: FADA \cite{Peng2020} and FedHarmony \cite{Dinsdale2022b}, both of which require the presence of the source data. Finally, we compare to SFDA methods: entropy minimisation; SHOT \cite{Liang2020}, USFAN \cite{Roy2022}, and gSFDA \cite{Yang2021}. We do not compare to any generative SFDA methods, as the ability to create source data would not meet privacy requirements for many applications \cite{Wang2022}, especially given that GANs often replicate training images when trained with small datasets \cite{Feng2021}. Details are provided in the Supplementary Material.

\textbf{Methods Comparison:} We first demonstrate the method for a range of batchsizes (5, 50 and 500) because methods that minimise entropy are expected to be more stable when using large batchsizes, which are rarely achievable when working with MR images due to the memory constraints posed by large image sizes \cite{Dinsdale2022}. Thus, robustness to the batchsize is vital if a SFDA method is to be used for harmonisation.  We use a single source model to allow fair comparison, trained with a batchsize of 50. We wish to maximise performance across all sites: as harmonisation is normally framed as a joint domain adaptation problem \cite{Dinsdale2021}, the average performance across all sites is reported. 

The results can be seen in Table \ref{tab: mednist results}, alongside the baseline methods. It can be seen that \texttt{SFHarmony} outperforms the existing SFDA methods, especially when a small batchsize was used for training (86.22\% for batchsize 5). SHOT \cite{Liang2020} showed comparable performance to \texttt{SFHarmony} when trained with a batchsize of 500 (85.27\%), but was  highly dependent on the modified source training. Interestingly, several of the SFDA approaches outperformed the adversarial approaches despite them having access to the source data, possibly due to the instability of such approaches. 

The proposed $D_{GMM}$ loss is clearly able to align the features across sites using only the GMM summary statistics. This is demonstrated by Fig. \ref{fig:features}, which shows the source and target features for each site before and after DA. Clearly the features overlap much more after DA, which both leads to the clear improvement in performance, and shows that the approach is achieving the harmonisation goals of the model having a shared feature embedding across sites. 

We tried modelling the features with $K\in\{1,2,3\}$ GMM components: visual inspection of the features suggested that at least 2 components would be beneficial. This was confirmed by the results,  with the best performance being achieved when modelling the features with 2 components, as shown in Table \ref{tab: mednist results}. However, the approach still performed well for 1 and 3 components, showing limited sensitivity to the number of components chosen. The number of components chosen is the only additional hyperparameter to be tuned with our approach, with only a single loss function to minimise. The results clearly show the robustness to the choice of batchsize, and the results were also robust to the choice of learning rate, with the accuracy staying within $1\%$ of the best result across learning rates from $10^{-7}$ to $10^{-4}$. Therefore, deployment of the proposed approach requires no changes for a new site, only the choice of the number of components for a new source model. This is in contrast to many existing SFDA approaches that require balancing several loss functions (e.g. \cite{Liang2020, Bateson2022}).

\textbf{Ablation Study:} We considered the GMM with $K=1$, allowing us to explore the w/o EM case, where $\bm{\mu}$ and $\bm{\sigma}^2$ are calculated directly. We also considered removing the batch memory across the training loop, reinitialising the EM algorithm before each batch. From  Table \ref{tab: mednist results} it is clear that both aspects are contributing to the performance, especially for small batchsizes. 

\begin{figure}
\centering
\includegraphics[width=0.45\textwidth]{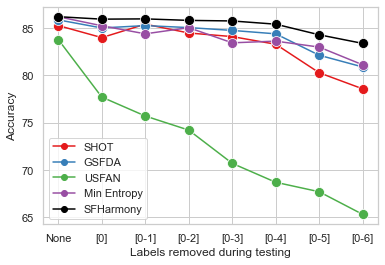}
\caption{Average accuracy across the sites with increasing numbers classes removed from the target site training, creating increasingly imbalanced data distributions. The x axis shows the classes that were removed. }
\label{fig:removelabels}
\end{figure}

\textbf{Class Imbalance:} In the above experiments, the distribution of class labels was approximately equal across sites. We now consider the extreme scenario where the source site contains samples from across all classes but target sites are missing classes. This is a conceivable scenario when considering MR images, where a given clinical site specialises in a certain condition and we are trying to harmonise the data to a carefully curated research dataset. Figure \ref{fig:removelabels} shows the average accuracy across sites, when the target sites had samples from an increasing number of classes removed. Each comparison method was trained using the best setting from Table \ref{tab: mednist results}. The proposed \texttt{SFHarmony} approach was more robust to the increased class imbalance than existing SFDA methods. 

\begin{figure}
\centering
\includegraphics[width=0.45\textwidth]{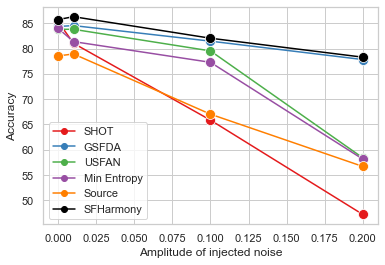}
\caption{Average accuracy across the sites with increasing magnitudes of noise injected into the source weights before communication. Amplitude is as a proportion of the source weights magnitude.}
\label{fig:differential}
\end{figure}

\textbf{Differential Privacy (DP):} Finally, we considered simulating the approach when DP is being used to further protect privacy. We simply simulated a Laplace mechanism of DP \cite{Dwork2014}, by injecting noise onto the weights before communication, modelled as: $\bm{w} = \bm{w} + Lap(|\bm{w}| f)$
where $f$ was varied to create increasing levels of noise. As the GMM is fit at the local site, $\bm{\Theta}_{Q^s}$ can be calculated before the noise is applied. The comparison methods were again all trained using the best setting from Table \ref{tab: mednist results}. Although this is a very simple model of DP, with many more sophisticated approaches existing, Fig. \ref{fig:differential} demonstrates that many existing methods for SFDA are very sensitive to the applied noise. SHOT \cite{Liang2020} is the most dramatically affected, with the pseudo-labelling approach suffering a significant degradation in performance. Our proposed approach maintained performance well across the applied noise levels, despite the frozen label predictor imposing a ceiling on performance. 

\begin{table*}[]
\resizebox{\textwidth}{!}{%
\begin{tabular}{|l|ccc|c|ccl|ccl|}
\hline
\multicolumn{1}{|c|}{\multirow{2}{*}{Method}} & \multicolumn{1}{c|}{\multirow{2}{*}{S}} & \multicolumn{1}{c|}{\multirow{2}{*}{T}} & \multirow{2}{*}{C} & \multirow{2}{*}{\begin{tabular}[c]{@{}c@{}}Information\\ Communicated\end{tabular}} & \multicolumn{3}{c|}{CC359 Average Dice} & \multicolumn{3}{c|}{ABIDE Average Dice} \\ \cline{6-11} 
\multicolumn{1}{|c|}{} & \multicolumn{1}{c|}{} & \multicolumn{1}{c|}{} &  &  & \multicolumn{1}{l|}{Bs 5} & \multicolumn{1}{l|}{Bs 50} & Bs 500 & \multicolumn{1}{l|}{Bs 5} & \multicolumn{1}{l|}{Bs 50} & Bs 500 \\ \hline
Source Model & $\checkmark$ & x & x & - &  & 0.832 &  &  & 0.775 &  \\ \hline
Centralised Training & $\checkmark$ & $\checkmark$ & $\checkmark$ & All Data & 0.983 & 0.985  & \multicolumn{1}{c|}{0.983} & 0.884 & 0.885 & \multicolumn{1}{c|}{0.875} \\
Target Finetune & x & $\checkmark$ & x & Model Weights & 0.981 & 0.982  & \multicolumn{1}{c|}{0.982} & 0.883 & 0.884 & \multicolumn{1}{c|}{0.885} \\ \hline
DeepCORAL \cite{Sun2016} & $\checkmark$ & x & $\checkmark$ & All Data & 0.768 & -  & \multicolumn{1}{c|}{-} & 0.523 & - & \multicolumn{1}{c|}{-} \\
FADA \cite{Peng2020} & $\checkmark$ & x & x & Model Weights + Features & 0.967 & 0.964  & \multicolumn{1}{c|}{0.959} & 0.830 & 0.827  & \multicolumn{1}{c|}{0.825} \\
FedHarmony \cite{Dinsdale2022b} & $\checkmark$ & x & x & Model Weights + Statistics & 0.965 & 0.962  & \multicolumn{1}{c|}{0.950} & 0.825 & 0.810 & \multicolumn{1}{c|}{0.822} \\ \hline
Minimise Entropy  & x & x & x & Model Weights & 0.767 & 0.849  & \multicolumn{1}{c|}{0.951} & 0.570 & 0.542 & \multicolumn{1}{c|}{0.659} \\
AdaEnt \cite{Bateson2020}  & x & x & x & Model Weights & 0.827 & 0.817  & \multicolumn{1}{c|}{0.962} & 0.625 & 0.656  & \multicolumn{1}{c|}{0.682} \\
AdaMI \cite{Bateson2022}  & x & x & x & Model Weights & 0.820 & 0.835  & \multicolumn{1}{c|}{0.965} &  0.606 &  0.657& \multicolumn{1}{c|}{0.660} \\
Direct Fit  & x & x & x & Model Weights + Statistics & 0.648 & 0.696  & \multicolumn{1}{c|}{0.873} & 0.615 & 0.803 & \multicolumn{1}{c|}{0.830} \\ \hline
SFHarmony 1 GMM Component  & x & x & x & Model Weights + Statistics & 0.950 & 0.949   & \multicolumn{1}{c|}{0.959} &  0.831 & $\bm{0.832}$ & \multicolumn{1}{c|}{0.831} \\
SFHarmony 2 GMM Components  &  x & x & x & Model Weights + Statistics & 0.970 & $\bm{0.970}$   & \multicolumn{1}{c|}{$\bm{0.970}$} &  0.832 & $\bm{0.832}$ & \multicolumn{1}{c|}{$\bm{0.832}$} \\
SFHarmony 3 GMM Components  & x & x & x & Model Weights + Statistics & $\bm{0.972}$ & 0.968   & \multicolumn{1}{c|}{$\bm{0.970}$} &  $\bm{0.833}$ & $\bm{0.832}$ & \multicolumn{1}{c|}{$\bm{0.832}$} \\ \hline

\end{tabular}%
}
\caption{Results on the CC359 dataset for brain extraction, and the ABIDE dataset for the tissue segmentation. S = Source data required, T = Target labels required, C = Centralised data, Bs = batchsize. The average Dice score is the performance across all 5 (CC359) /4 (ABIDE) sites, weighted equally, and is reported for training batchsizes of 5, 50 and 500. The best performing SFDA method for each batchsize for each segmentation task is in bold.}
\label{tab: segmentation}
\end{table*}

\begin{table*}[]
\centering
\resizebox{0.75\textwidth}{!}{%
\begin{tabular}{|l|ccc|c|cll|}
\hline
\multicolumn{1}{|c|}{\multirow{2}{*}{Method}} & \multicolumn{1}{c|}{\multirow{2}{*}{S}} & \multicolumn{1}{c|}{\multirow{2}{*}{T}} & \multirow{2}{*}{C} & \multirow{2}{*}{\begin{tabular}[c]{@{}c@{}}Information\\ Communicated\end{tabular}} & \multicolumn{3}{c|}{Average MAE} \\ \cline{6-8} 
\multicolumn{1}{|c|}{} & \multicolumn{1}{c|}{} & \multicolumn{1}{c|}{} &  &  & \multicolumn{1}{c|}{Bs 4} & \multicolumn{1}{c|}{Bs 8} & Bs 16 \\ \hline
Source Model & $\checkmark$ & x & x & - &  & \multicolumn{1}{c}{4.38} &  \\ \hline
Centralised Training & $\checkmark$ & $\checkmark$ & $\checkmark$ & All Data & 3.52 & \multicolumn{1}{c}{3.38} & \multicolumn{1}{c|}{3.36} \\
Target Finetune & x & $\checkmark$ & x & Model Weights & 3.57 & \multicolumn{1}{c}{3.60} & \multicolumn{1}{c|}{3.58} \\ \hline
DeepCORAL \cite{Sun2016} & $\checkmark$ & x & $\checkmark$ & All Data & 4.58 & \multicolumn{1}{c}{4.41} & \multicolumn{1}{c|}{4.12} \\
FADA \cite{Peng2020} & $\checkmark$ & x & x & Model Weights + Features & 3.55 & \multicolumn{1}{c}{3.42} & \multicolumn{1}{c|}{3.78} \\
FedHarmony \cite{Dinsdale2022b} & $\checkmark$ & x & x & Model Weights + Statistics & 3.61 & \multicolumn{1}{c}{3.50} & \multicolumn{1}{c|}{3.79} \\
Direct Fit & x & x & x & Model Weights + Statistics & 4.70 & \multicolumn{1}{c}{4.31} & \multicolumn{1}{c|}{4.05} \\ \hline
SFHarmony 1 GMM Component & x & x & x & Model Weights + Statistics & 4.21 & \multicolumn{1}{c}{4.13} & \multicolumn{1}{c|}{3.71} \\
 SFHarmony 2 GMM Components & x & x & x & Model Weights + Statistics & 3.87 & \multicolumn{1}{c}{$\bm{3.72}$} & \multicolumn{1}{c|}{$\bm{3.69}$} \\
  SFHarmony 3 GMM Components & x & x & x & Model Weights + Statistics & $\bm{3.64}$ & \multicolumn{1}{c}{$\bm{3.72}$} & \multicolumn{1}{c|}{3.73} \\
  \hline
  
\end{tabular}%
}
\caption{Results on the ABIDE dataset for the age prediction task. S = Source data required, T = Target labels required, C = Centralised data, Bs = Batchsize. The average MAE is the performance across all 4 sites, weighted equally, and is reported for training batchsizes of 4, 8 and 16: 16 was the largest batch achievable. The best SFDA method for each batchsize is in bold.}
\label{tab: regression}
\end{table*}

\subsection{Segmentation: CC359 and ABIDE datasets}
We now demonstrate our approach on two multisite MRI datasets for segmentation tasks: brain extraction (CC359) with two labels (brain/background) and tissue segmentation (ABIDE) with four labels (WM, GM, CSF, background). 

\textbf{Baselines:} There are far fewer existing methods for SFDA, and we again did not compare to generative approaches. Thus, we compared to supervised oracles: source model only, centralised data and target finetuning with frozen label predictor; semisupervised approaches: DeepCORAL \cite{Sun2016}, FADA \cite{Peng2020} and FedHarmony \cite{Dinsdale2022b}; then for SFDA approaches we compared to minimising entropy, AdaEnt \cite{Bateson2020} and AdaMI \cite{Bateson2022}, and Direct Fit (w/o EM in ablation study). We were unable to train DeepCORAL with a batchsize of more than 5 due to memory constraints.

\textbf{Methods Comparison:} Table \ref{tab: segmentation} shows the results for both tasks. In the classification task there were only $32$ features in the fully connected layer; however, now there are many more, for instance for the CC359 data there are $65536$ features across all of the convolutional filters. Despite this increase in features, \texttt{SFHarmony} was able to complete the DA for both segmentation tasks, leading to an improved Dice score over the existing methods, across the batchsizes considered. Again, the existing SFDA methods were very sensitive to batchsize, and AdaMI \cite{Bateson2022} was also sensitive to the choice of tissue ratio prior: as we were completing the segmentation tasks on 2D slices, different slices had varying amounts of the target label present and we had to create a prior that was dependent on slice depth to achieve reasonable performance. The ABIDE tissue segmentation task was more challenging, as can be seen by the comparatively lower Dice Scores, especially due to the large imbalance in tissues, which affected the performance of AdaMI. 

No changes needed to be made to the approach compared to the classification task, including the learning rate, showing the generalisability of the method across tasks. 

\subsection{Regression: ABIDE dataset}
\textbf{Baselines:} We could not identify any appropriate SFDA baselines. Therefore, the only comparison methods were: source model only, centralised data and target finetuning with frozen label predictor, DeepCORAL \cite{Sun2016}, FADA \cite{Peng2020}, FedHarmony \cite{Dinsdale2022b} and Direct Fit. The maximum batchsize possible was 16, and so we tried batchsizes of 4, 8 and 16.

\textbf{Methods Comparison:} 
It can be seen from Table \ref{tab: regression} that for the age prediction task FADA \cite{Peng2020} outperformed our proposed approach for two of the three reported batchsizes, unlike in the other tasks. This may well be because the task was completed in 3D, and thus a small number of samples were available, meaning that the presence of the source data supported the model training. \texttt{SFHarmony} did, however, show comparable performance, especially when modelling the features with more components. We could not identify any SFDA methods in the literature that could be directly applied to regression tasks. Our method is flexible and can be directly applied to the regression task without any change to the model architecture or DA procedure. 

\section{Conclusion}
We have presented \texttt{SFHarmony}, a method for SFDA, motivated by the need to harmonise MRI data across imaging sites while relaxing assumptions about the availability of source data. We have demonstrated the applicability of the method to classification, regression, and segmentation tasks, and have shown that it outperforms existing SFDA approaches when applied to MR imaging data. The approach is general, allowing it to be applied across architectures and tasks. Issues may arise due the increase in features when applying the approach to 3D volumes. Currently, the approach models each feature as an independent GMM, but features will be highly related within a filter and approaches to utilise these relations should be explored.

\section{Acknowledgements}
ND is supported by a Academy of Medical Sciences Springboard Award. MJ
is supported by the National Institute for Health Research, Oxford Biomedical Research Centre, and this research was funded by the Wellcome Trust
[215573/Z/19/Z]. WIN is supported by core funding from the Wellcome Trust
[203139/Z/16/Z]. AN is grateful for support from the Academy of Medical Sciences under the Springboard Awards scheme (SBF005/1136), and the Bill and Melinda Gates Foundation.

{\small
\bibliographystyle{ieee_fullname}
\bibliography{references}
}

\end{document}